\documentclass[journal]{IEEEtran}
\usepackage{epsfig}
\usepackage{graphicx}
\usepackage{amsmath}
\usepackage{amssymb}
\usepackage{amsfonts}
\usepackage{CJK}
\usepackage{amsmath}
\usepackage{algorithm}
\usepackage{algorithmic}
\usepackage{bm}
\usepackage{indentfirst}
\usepackage{multirow}
\usepackage{graphicx}
\usepackage{array}
\newcolumntype{C}[1]{>{\centering}p{#1}}
\ifCLASSINFOpdf
\else
\fi
\begin{document}
%
\title{Multi-modal Fusion for Diabetes Mellitus and Impaired Glucose Regulation Detection}
%
%
%

\author{Jinxing Li,
        David Zhang,~\IEEEmembership{Fellow,~IEEE,}
        Yongcheng Li, and Jian Wu
\thanks{J. Li is with the Department
of Computing, Hong Kong Polytechnic University, Hung Hom, Kowloon (e-mail: csjxli@comp.polyu.edu.hk).}
\thanks{D. Zhang is with the Department
of Computing, Hong Kong Polytechnic University, Hung Hom, Kowloon (e-mail: csdzhang@comp.polyu.edu.hk).}
\thanks{Y. Li is with the Department of Computer Science£¬Harbin Institute of Technology Shenzhen graduate school, Shenzhen, China (email: liyongchengem@126.com).}
\thanks{J. Wu is with the Department of Computer Science£¬Harbin Institute of Technology Shenzhen graduate school, Shenzhen, China (email: wujianhitsz@gmail.com).}

\thanks{Manuscript received XXX; revised XXX.}}

%
%

\markboth{Journal of XX Class Files,~Vol.~XX, No.~X, XX~2016}%
{Shell \MakeLowercase{\textit{et al.}}: Bare Demo of IEEEtran.cls for IEEE Journals}
%



\maketitle

\begin{abstract}
Effective and accurate diagnosis of Diabetes Mellitus (DM), as well as its early stage Impaired Glucose Regulation (IGR), has attracted much attention recently. Traditional Chinese Medicine (TCM) [3], [5] etc. has proved that tongue, face and sublingual diagnosis as a noninvasive method is a reasonable way for disease detection. However, most previous works only focus on a single modality (tongue, face or sublingual) for diagnosis, although different modalities may provide complementary information for the diagnosis of DM and IGR. In this paper, we propose a novel multi-modal classification method to discriminate between DM (or IGR) and healthy controls. Specially, the tongue, facial and sublingual images are first collected by using a non-invasive capture device. The color, texture and geometry features of these three types of images are then extracted, respectively. Finally, our so-called multi-modal similar and specific learning (MMSSL) approach is proposed to combine features of tongue, face and sublingual, which not only exploits the correlation but also extracts individual components among them. Experimental results on a dataset consisting of 192 Healthy, 198 DM and 114 IGR samples (all samples were obtained from Guangdong Provincial Hospital of Traditional Chinese Medicine) substantiate the effectiveness and superiority of our proposed method for the diagnosis of DM and IGR, compared to the case of using a single modality.
\end{abstract}


%
\IEEEpeerreviewmaketitle

\section{Introduction}
%
%
%
%
\IEEEPARstart{T}{he} number of people suffering from diabetes mellitus (DM) is increasing each year, which is predicted to reach 366 million by 2030 [1], causing disabilities, economic hardship and even death. An accurate diagnosis of DM, especially for its early stage also known as Impaired Glucose Regulation (IGR), is becoming more and more important. Until now, the fasting plasma glucose (FPG) test is a standard method to diagnose DM in many hospitals. FPG test is performed by analyzing the patient's blood glucose level after the patient has gone at least 12 hours without taking any food. This method is accurate, but inconvenient and painless. This blood required detecting method can be consider invasive and  slightly painful, and even has a risk of infection (piercing process).

In recent years, some works have been done on non-invasive methods to diagnose specific diseases by using body surface features (the tongue, face, and sublingual vein). The human tongue, face and sublingual vein contain numerous valuable information that can be used for diagnosis [12], [13], [14], [15], [16], [17] with color, texture and geometry features being the most prominent.

\begin{figure}
  \centering
  \includegraphics[width=80mm]{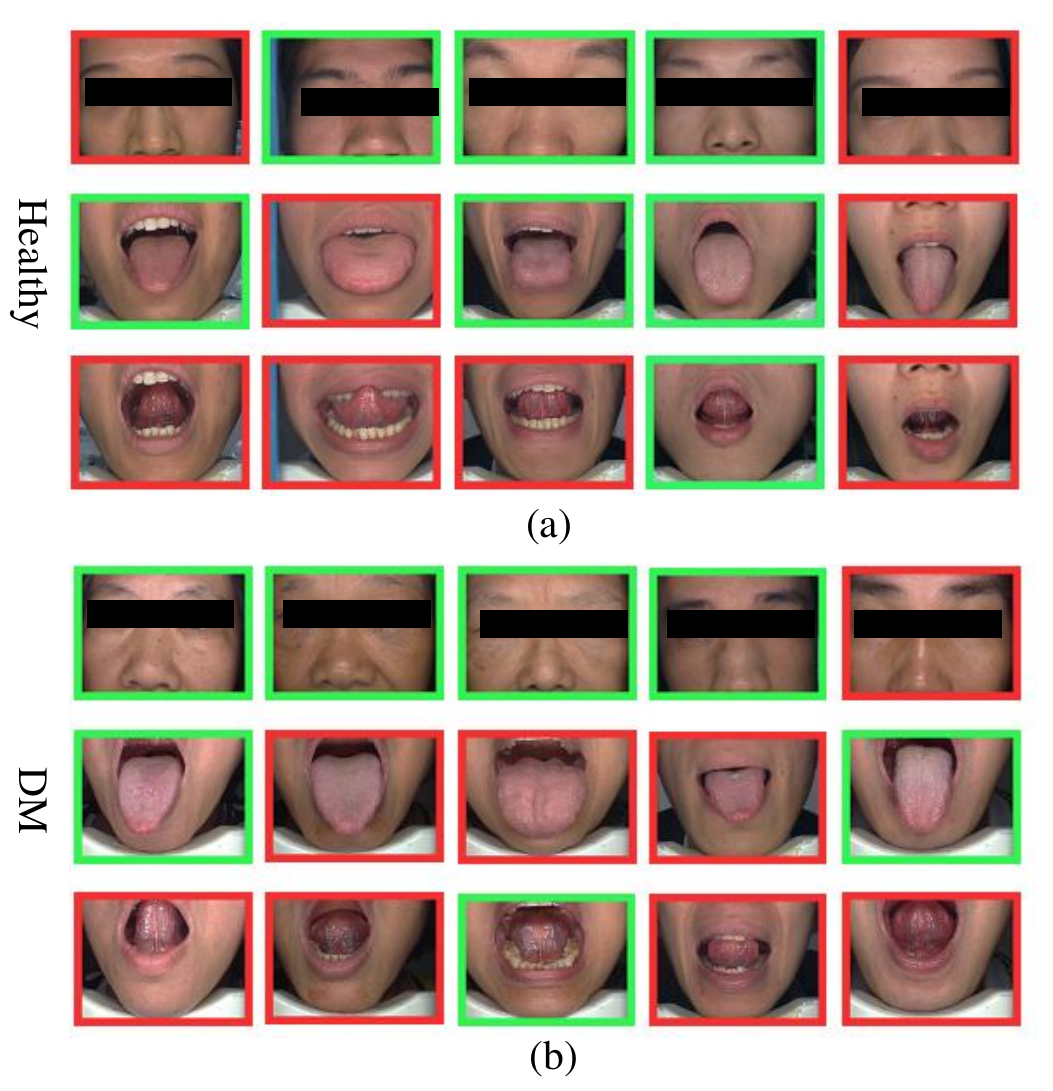}
  \caption{Example classifications using Sparse Representation Classification. A green border indicates correct classification and a red border represents incorrect classification. The first row represents experimental results using facial images, followed by the tongue images and sublingual images in the second row and third row of the 'Healthy' part. So does the 'DM' part.}
  \label{fig:env}
\end{figure}

The experimental results of disease diagnosis based on tongue, facial and sublingual images have proved the effectiveness and reasonability of aforementioned non-invasive methods. [2] first captured a precise facial image for diagnosis using a chamber with LED light and digital camera. They established a five color scale for facial image to measure the changes caused by internal organs. Wang et al. [2] proposed a mathematically described tongue color space. They statistically studied main 12 types of color distribution in tongue with over 9000 tongue images, and their corresponding experiments illustrated that these colors contributed to the disease classification. In [3], Zhang et al. used facial block color with sparse representation classifier (SRC) [4] for DM detection. Moreover, extracted color, texture and geometry features of tongue feature were exploited to detect DM and nonproliferative diabetic retinopathy (DR) [5]. A heart disease diagnostic system based on facial color [18] was proposed. Five facial blocks were extracted from the facial image to detect hepatitis in [19], and the average accuracy achieved 73.6\% using the average RGB pixel intensities features. Zhang and Wang etc. [23] took both color and texture features into account for computerized facial diagnosis.

However, despite of various tongue, face or sublingual diagnosis methods proposed for disease detection, most of them regarded either tongue, face or sublingual vein as an independent one and ignored the relationship among them which may have an effect on the overall classification performance. As shown in Fig. 1, it is easy to see that some healthy or DM samples can not be classified with tongue features but facial features or sublingual features do. Similarly, some samples can be detected by the tongue task or the sublingual task but the facial task is unable. In particular, some patients are difficult to be diagnosed with tongue, face and sublingual vein (e.g., the last column in 'Healthy' part of Fig. 1), while a combination of these tasks may have a possibility for accurate diagnosis. Thus, an effective exploitation of the complementary information is beneficial for the diagnosis of diseases. A naive way of taking the tongue, face and sublingual vein into account is to concatenate these three task vectors as a single one. However, it is not an efficient way since these three modalities are different. Furthermore, the concatenated feature dose not exploit the cross correlated information among the original data. Therefore, it is necessary to have a research in modal combination with the tongue, facial and sublingual tasks.

In this paper, we propose a novel multi-modal classifier to discriminate between DM (or IGR) and healthy controls. In particular, our proposed method jointly represents three modal features obtained from tongue, facial and sublingual images and shares a similarity between them. In addition, consider differences between those tasks which contain useful information for classification, we also extract individual components that keeps the diversity between them. In this case, both similarity and distinctiveness of multiple modalities are exploited,  being beneficial for the disease detection. An optimal algorithm based on Linearized Alternating Directions Method (LADM) [24] and Augmented Lagrangian Multiplier (ALM) method [6] is applied to solve the presented strategy.

The rest of this paper is organized as follows. In Section 2, we briefly describe our previous work including the image capture device and the corresponding feature extraction. In Section 3, we analyzes the proposed multi-modal classifier. Section 4 illustrates the experimental results, followed by concluding remarks in Section 5.
\section{Image Capture Device and Feature Extraction}
In this section, we will first introduce the image capture device of the tongue, face and sublingual vein, and then describe the feature extraction of these three types of images.
\subsection{Tongue and Facial Capture Device}

\begin{figure}
  \centering
  \includegraphics[width=90mm]{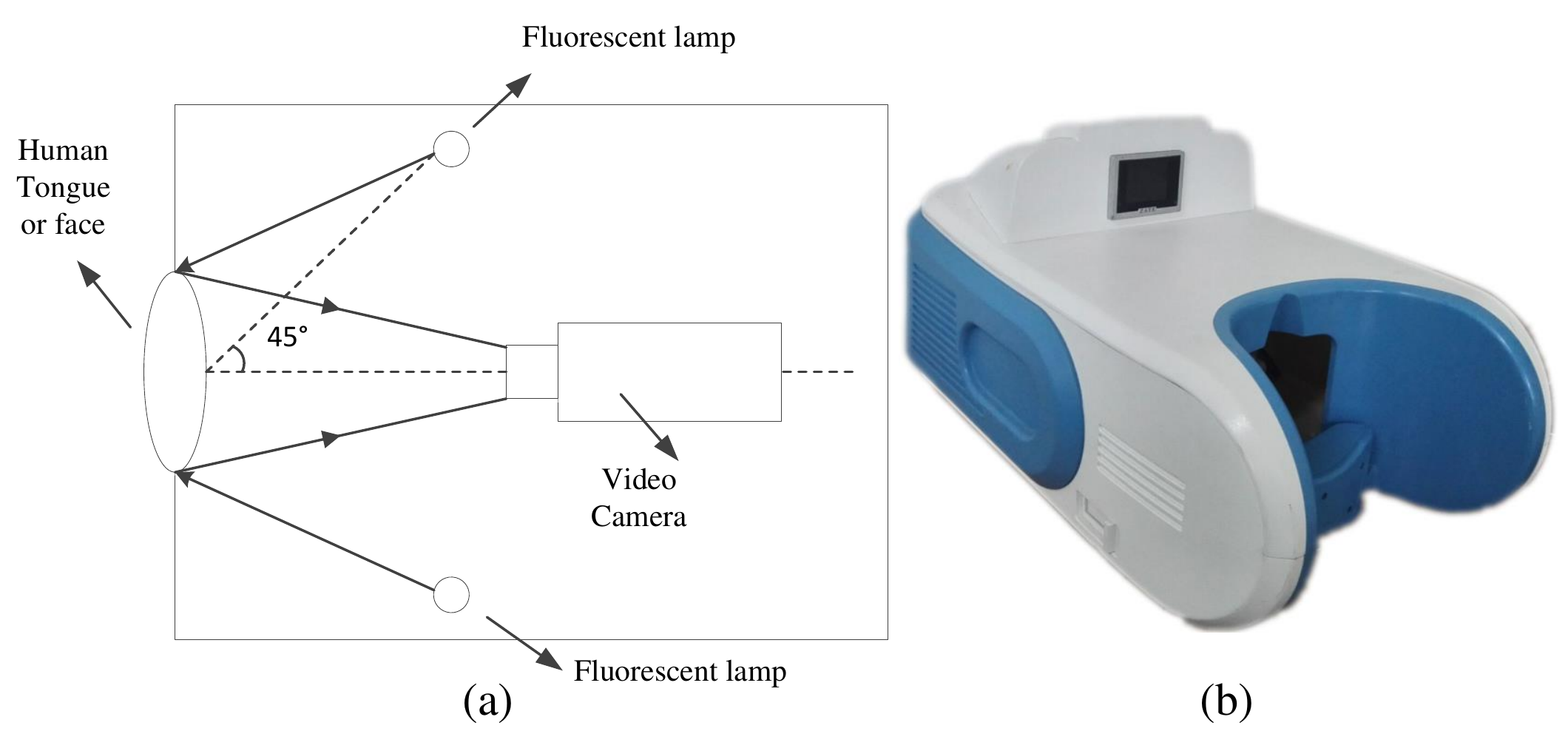}
  \caption{Image capture device. (a) Viewing geometry and imaging path of the imaging device. (b) Appearance of the device and the system.}
  \label{fig:env}
\end{figure}

The image capture (tongue, face and sublingual) device is shown in Fig. 2. consisting of a SONY 3-CCD video camera placed in the center and two D65 fluorescent tubes situated symmetrically on either sides of the camera in order to produce a uniform illumination. Particularly, the angle between the incident light and emergent light is 45 (shown in Fig. 2(a)), recommended by Commission Internationale de l'Eclairage (CIE). For the tongue and sublingual image capture, patients placed their chin on a chin rest and show their tongue to the camera (shown in Fig. 2(b)), while for the facial image capture, patient placed their chin on a chin rest and show their face to the camera (change the height or position of the chin rest to obtain the tongue, facial and sublingual images). Each image saved in JPEG format with 640$\times $480 size is color corrected [20] to eliminate any variability in color images caused by changes of the illumination and device dependence. Using this correction model, original tongue, facial and sublingual images are corrected, and pixels are transformed in the standard RGB (sRGB) color space.
\subsection{Feature Extraction}
In our previous work [3], [5], we have proposed a method to extract color, texture and geometry features of the three types of images. In this subsection, we will briefly introduce them. It should be noted that refer to [21], four blocks with 64$\times$64 size strategically located around the face are extracted, which contain the information of the health status of the human. Details of the location of blocks can be found in [30]. Both color and texture extraction for facial images are based on these blocks. Similarly, we also have defined eight blocks with 64$\times $64 size for tongue texture feature extraction. More details can be found in [5]. For the sublingual images, the main domain about the vein is first decomposed and its color and geometry features are then extracted.
\subsubsection{Color Feature}

The RGB values of captured tongue or facial images are first calculated and then converted to CIEXYZ
\begin{equation}
\begin{bmatrix}
X\\
Y\\
Z
\end{bmatrix}=\begin{bmatrix}
0.4124 &0.3576  &0.1805 \\
0.2126 &0.7152  &0.0722 \\
0.0193 &0.1192  &0.9505
\end{bmatrix}\begin{bmatrix}
R\\
G\\
B
\end{bmatrix}
\end{equation}
followed by CIEXYZ to CIELAB \cite{zhang2005svr}
\begin{equation}
\begin{aligned}
&L=166f(Y/Y_{0})-16\\
&a=500[f(X/X_{0})-f(Y/Y_0)]\\
&b=200[f(Y/Y_{0})-f(Z/Z_{0})]
\end{aligned}
\end{equation}
where $X_{0}$, $Y_{0}$ and $Z_{0}$ are the CIEXYZ tristimulus values of the reference white point; $f(x)=x^{1/3}$ if $x>0.008856$ or $f(x)=7.787x+16/116$ if $x\leq  0.008856$.

We then compare the obtained LAB values with 12 predefined colors for the tongue [5], 6 predefined colors for the face [3] and 6 predefined colors (the method of defining these colors is same in the three types of images) for the sublingual vein to assign the color value which is closest to it using Euclidean distance. After evaluating all tongue, facial or sublingual pixels, the total of each color is summed and divided by the total number of pixels. We regard these ratios as the color feature. For tongue image, we statistically extract 12 different colors as the feature. Thus, a 12 dimensional vector for the tongue is obtained. Similarly, we select 6 different colors and 6 different colors for face and sublingual (more detailed color feature extraction for sublingual images will be published soon), respectively. Finally, a 6 dimensional vector for each block of the face and a 6 dimensional vector for the sublingual vein are acquired.

\subsubsection{Texture Feature}
The 2-D Gabor filter is applied to calculate the texture of each block.
\begin{equation}
G_{k}(x,y)=\exp\left ( \frac{x^{'2}+\gamma ^{2}y^{'2}}{-2\sigma ^{2}} \right )\cos\left ( 2\pi \frac{x'}{\lambda } \right )
\end{equation}
where $x'=x\cos\theta+y\sin\theta$, $y'=-x\sin\theta+y$, $\theta$ is the orientation,  $\gamma$ is the aspect ratio of the sinusoidal function, $\sigma$ is the variance, and $\lambda$ is the wavelength. A response $R_{k}(x,y)$ is produced by convolving each filter with a texture block.
\begin{equation}
R_{k}(x,y)=G_{k}(x,y)* im(x,y)
\end{equation}
where the symbol $*$ denotes 2-D convolution and the function $im(x,y)$ represents the texture block. Then the maximum pixel intensity is selected following $FR(x,y)=\max(R_{1}(x,y), \cdots  ,R_n(x,y))$. Apart from the texture value of each block, we also add the mean of these values as an additional texture value. Thus, we can get a 9 dimensional vector for the tongue and 5 dimensional vector for each block of the face.

\subsubsection{Geometry Feature}
Statistically, a person who suffers from DM or IGR would affect the geometry of the tongue [5] and the sublingual vein [25]. In our previous work, we have introduced the details of the geometry feature extraction [5] and these features are based on measurements, distances, areas, and their ratios. For tongue images, 13 geometry features are selected (width, length, length-width ratio, smaller half-distance, center distance, center distance ratio, area, circle area, circle area ratio, square area, square area ratio, triangle area, and triangle area ratio). For sublingual images (more detailed geometry feature extraction for sublingual images will be published soon), 6 geometry features are selected (length, width, length ratio of each side vein). Note that, before extracting the geometry feature of sublingual images, we first decompose the image to get the sublingual vein.

\section{Multi-Modal Similar and Specific Learning}
A general framework for multi-modal fusion is proposed in this section. Before introducing our proposed method, we first briefly review the sparse representation classifier (SRC).


\subsection{Sparse Representation Classifier}
Given a set of training samples and a test sample, the main idea
of SRC is that the test sample is represented as a linear combination
of the training samples, and the representation coefficients are required to be as sparse as possible. In practice, the $l_{1}$-norm minimization is applied to ensure the sparsest linear
representation of the test sample over the training samples. Suppose that matrix $\mathbf{D}=[\mathbf{D}_{1},\mathbf{D}_{2},\cdot ,\cdot ,\cdot ,\mathbf{D}_{J}]$ called dictionary is the set of training samples, where $J$ is the number of classes, $\mathbf{D}_{i}\in \mathbb{R}^{m\times n_{i}}$ is the training set of the $i$-$th$ class with $m$ dimension and $n_{i}$ samples. Each column in $\mathbf{D}$ is called atom. A test sample $\mathbf{y}$ can be denoted by
\begin{equation}
\hat{\boldsymbol{\alpha }}=\min_{\boldsymbol{\alpha }}\left \| \mathbf{y }-\mathbf{D }\boldsymbol{\alpha } \right \|_{F}^{2}+\lambda \left \| \boldsymbol{\alpha } \right \|_{1}
\end{equation}
where $\lambda$ is the penalty parameter, $\left \| \cdot  \right \|_{F}$ is the Frobenius norm and $\left \| \cdot  \right \|_{1}$ is $\l _{1}$ norm. $\hat{\boldsymbol{\alpha }}=[\hat{\boldsymbol{\alpha }}_{1};\hat{\boldsymbol{\alpha }}_{2};\cdot \cdot \cdot ;\hat{\boldsymbol{\alpha }}_{J}]$ is the sparse coefficient, and $\hat{\boldsymbol{\alpha }}_{i}$ is the sparse coefficient corresponding to $\mathbf{D}_{i}$.

If the test sample $\mathbf{y}$ is from
class $i$, then it can be well represented by the training samples from $i$-$th$ class. In other words, among its representation coefficients $\hat{\boldsymbol{\alpha }}$ over all the
training samples, only coefficients in class $i$ will
be significant while others will be insignificant. Then the class label of the test sample is determined by the following Eq. (6)

\begin{equation}
i^{*}=\min\left \| \mathbf{y}-\mathbf{D}_{i}\hat{\boldsymbol{\alpha }}_{i} \right \|_{F}^{2}
\end{equation}

More information about the SRC can be found in [4].

\subsection{Multi-Modal Similar and Specific Learning Model}
\begin{figure*}
  \centering
  \includegraphics[width=140mm]{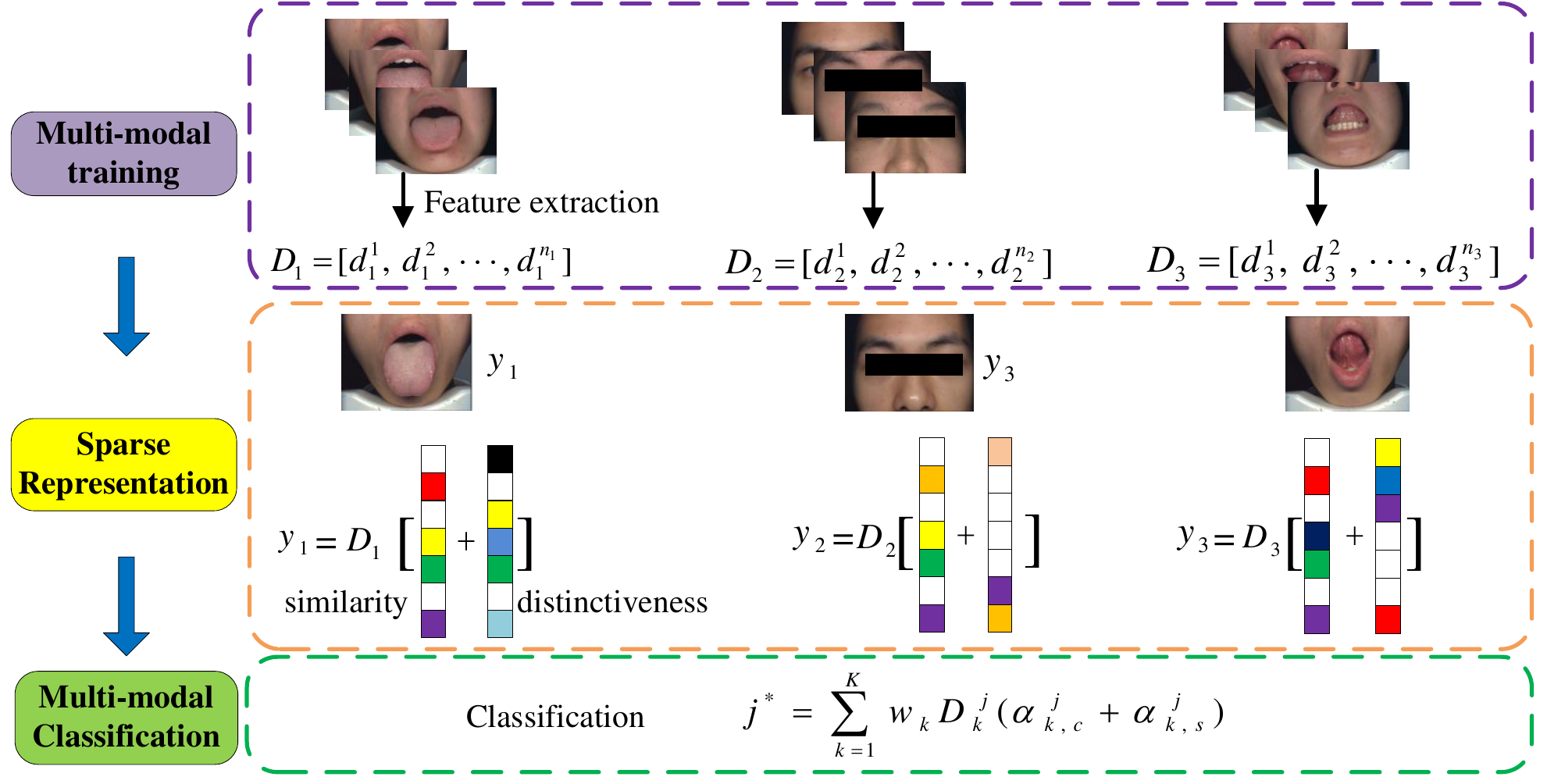}
  \caption{The framework of our proposed method MMSSL. MMSSL contains three parts: multi-modal training, sparse representation and multi-modal classification. Firstly, the dictionary consists of training samples; secondly, multi-modal features of a given test sample is represented sparsely with the dictionary, and the representation coefficients are divided into two parts which are common components and individual components; thirdly, the label of the test sample is decided according to the total reconstruction error.}
  \label{fig:env}
\end{figure*}

As mentioned above, different modal or feature vectors from a same sample may share
some similarity. It is reasonable to assume that representation coefficients coded on their associated dictionaries of different modalities should be similar which would make the
representation stable. For example, a test sample, containing tongue modality, facial modality and sublingual modality, is represented as a linear combination of the training samples; since all tasks are belonging to a same sample, they will be well represented by the training samples of their corresponding class, and hence the position and values of significant coefficients are similar. To achieve the above goal, we use [7] the following term to achieve the similarity of different modalities .
 \begin{equation}
 \min _{\boldsymbol{\alpha}_{k} }\sum_{k=1}^{K}\left \| \boldsymbol{\alpha} _{k}-\bar{\boldsymbol{\alpha} } \right \|_{F}^{2}
 \end{equation}
where $\boldsymbol{\alpha} _{k}$ is the representation coefficient of the $k$-$th$ modality and $\bar{\boldsymbol{\alpha} }=\frac{1}{K}\sum_{k=1}^{K}\boldsymbol{\alpha}_{k}$ is the mean vector of all $\boldsymbol{\alpha} _{k}$
($K$ is the number of different types of modalities). It is easy to see that Eq. (7) aims to reduce the variance of different representation coefficients $\boldsymbol{\alpha} _{k}$, making them similar to each one. However, this assumption is too restrictive since there is also distinctiveness among them. Therefore,
it is necessary to not only exploit the similarity among all
tasks but also keep the flexibility of each task. In this case,
the balance between similarity and distinctiveness among
all tasks will represent the original sample more stable
and accurate which is beneficial for classification.

To address aforementioned problem, we divide the representation coefficients $\boldsymbol{\alpha} _{k}$ into two parts: the similar part and the modal-specific part. Specifically, $\boldsymbol{\alpha} _{k}=\boldsymbol{\alpha} _{k}^{c}+\boldsymbol{\alpha} _{k}^{s}$, where $\boldsymbol{\alpha} _{k}^{c}$ denotes the similarity, while $\boldsymbol{\alpha} _{k}^{s}$ denotes the distinctiveness. The framework of our proposed method is shown in Fig. 3. The formulation of our model is
\begin{multline}
\min \sum_{k=1}^{K}\left \{  \left \| \mathbf{y}_{k}-\mathbf{D}_{k}\left ( \boldsymbol{\alpha}_{k}^{c}+\boldsymbol{\alpha}_{k}^{s} \right ) \right \|_{F}^{2}+\tau \left \| \boldsymbol{\alpha}_{k}^{c}-\bar{\boldsymbol{\alpha}}^{c} \right \|_{F}^{2}\right \}\\
+\sum_{k=1}^{K}\lambda \left ( \left \| \boldsymbol{\alpha}_{k}^{c} \right \|_{1}+\left \| \boldsymbol{\alpha}_{k}^{s} \right \|_{1} \right )
\end{multline}
where $\mathbf{y}_{k}$ is the test sample, $ \mathbf{D}_{k}=[\mathbf{D}_{k}^{1},\mathbf{D}_{k}^{2},\cdot ,\cdot ,\cdot ,\mathbf{D}_{k}^{J}]$ is training samples of the $k$-$th$ task, and $\mathbf{D}_{k}^{i}\in \mathbb{R}^{m_{k}\times n_{k}^{i}}$ is the training set of the $k$-$th$ task of the $i$-$th$ class with $m_{k}$ dimension and $n_{k}^{i}$ samples; $\bar{\boldsymbol{\alpha}}^{c}=\frac{1}{K}\sum_{k=1}^{K}\boldsymbol{\alpha}_{k}^{c}$ is the mean value of similar sparse representation coefficients of all tasks, and $\tau$ and $\lambda$ are the non-negative penalty constants. From Eq. (8), we can see that our model aims to extract the similar components of each task through $\mathbf{x}_{k}^{c}$, while also keeps individual components of each task through $\mathbf{x}_{k}^{s}$. Note that at the step of exploiting correlation, we do not directly set each $\boldsymbol{\alpha}_{k}^{c}$ to be equal, but instead of minimizing the distance between them. This way also makes our approach more flexible. In addition, consider the test sample would be linearly represented by atoms of the dictionary belonging to its own class, we apply $l_{1}$ norm on both $\boldsymbol{\alpha}_{k}^{c}$ and $\boldsymbol{\alpha}_{k}^{s}$.
\subsection{Optimization of MMSSL}
 We alternatively update the similar coefficients $\boldsymbol{\alpha}_{k}^{c}$ and special coefficient $\boldsymbol{\alpha}_{k}^{s}$. For example, we update $\boldsymbol{\alpha}_{k}^{c}$ by fixing $\boldsymbol{\alpha}_{k}^{s}$, and vice versa.

\textbf{Update $\boldsymbol{\alpha}_{k}^{c}$:}
By fixing $\boldsymbol{\alpha}_{k}^{s}$, the optimization solution of Eq. (8) with respect to $\boldsymbol{\alpha}_{k}^{c}$ equals to the following problem
\begin{multline}
\boldsymbol{\alpha}_{k}^{c}=\arg\min\left \| \mathbf{y}_{k}-\mathbf{D}_{k}\left ( \boldsymbol{\alpha}_{k}^{c}+\boldsymbol{\alpha}_{k}^{s} \right ) \right \|_{F}^{2}+\tau \left \| \boldsymbol{\alpha}_{k}^{c}-\bar{\boldsymbol{\alpha}}^{c} \right \|_{F}^{2}\\+\lambda  \left \| \boldsymbol{\alpha}_{k}^{c} \right \|_{1}
\end{multline}
we apply Augmented Lagrangian method (ALM) algorithm to modify Eq. (9).

Applying the ALM, the problem of (9) can be modified as follows.
\begin{multline}
\boldsymbol{\alpha}_{k}^{c}=\arg\min\left \| \mathbf{y}_{k}-\mathbf{D}_{k}\left ( \boldsymbol{\alpha}_{k}^{c}+\boldsymbol{\alpha}_{k}^{s} \right ) \right \|_{F}^{2}+\tau \left \| \boldsymbol{\alpha}_{k}^{c}-\bar{\boldsymbol{\alpha}}^{c} \right \|_{F}^{2}\\+ \lambda \left \| {\boldsymbol{\alpha}_{k}^{c}}' \right \|_{1}
+\frac{\mu}{2} \left \| \boldsymbol{\alpha}_{k}^{c}-{\boldsymbol{\alpha}_{k}^{c}}'+\frac{\mathbf{z}_{k}}{\mu } \right \|_{F}^{2}
\end{multline}
where ${\boldsymbol{\alpha}_{k}^{c}}'$ is the relaxed variable, $\mathbf{z}_{k}$ is the $k$-$th$ lagrangian multiplier, and $\mu$ is the step value. Then we can optimize $\boldsymbol{\alpha}_{k}^{c}$ and ${\boldsymbol{\alpha}_{k}^{c}}'$ alternatively.

(a) Firstly, we fix ${\boldsymbol{\alpha}_{k}^{c}}'$ to get ${\boldsymbol{\alpha}_{k}^{c}}$
\begin{multline}
\boldsymbol{\alpha}_{k}^{c}=\arg\min \left \| \mathbf{y}_{k}-\mathbf{D}_{k}\left ( \boldsymbol{\alpha}_{k}^{c}+\boldsymbol{\alpha}_{k}^{s} \right ) \right \|_{F}^{2}+\tau \left \| \boldsymbol{\alpha}_{k}^{c}-\bar{\boldsymbol{\alpha}}^{c} \right \|_{F}^{2}\\
+\frac{\mu}{2} \left \| \boldsymbol{\alpha}_{k}^{c}-{\boldsymbol{\alpha}_{k}^{c}}'+\frac{\mathbf{z}_{k}}{\mu } \right \|_{F}^{2}
\end{multline}
Follow the Ref. [7], a closed-form solution of $\boldsymbol{\alpha}_{k}^{c}$ can be derived:
\begin{equation}
\boldsymbol{\alpha}_{k}^{c}=\boldsymbol{\alpha}_{0,k}^{c}+\frac{\tau }{K}\mathbf{P}_{k}\mathbf{Q}\sum_{\eta =1}^{K}\boldsymbol{\alpha}_{0,\eta }^{c}
\end{equation}
where $ \mathbf{P}_{k}=(\mathbf{D}_{k}^{T}\mathbf{D}_{k}+(\tau +\frac{\mu}{2})\mathbf{I})^{-1}$, $\boldsymbol{\alpha}_{0,k}^{c}=\mathbf{P}_{k}(\mathbf{D}_{k}^{T}(\mathbf{y}_{k}-\mathbf{D}_{k}\boldsymbol{\alpha}_{k}^{s})+\frac{\mu}{2}\boldsymbol{\alpha}_{k}^{c'}-\frac{\mathbf{z}_{k}}{2})$, and $\mathbf{Q}=(\mathbf{I}-\frac{\tau}{K}\sum_{\eta =1}^{K}\mathbf{P}_{\eta })^{-1}$.

(b) Secondly, after fixing ${\boldsymbol{\alpha}_{k}^{c}}$, the optimization solution of Eq. (10) can be reduced to Eq. (13) at the step of updating ${\boldsymbol{\alpha}_{k}^{c}}'$.
\begin{equation}
{\boldsymbol{\alpha}_{k}^{c}}'=\arg \min  \lambda  \left \| {\boldsymbol{\alpha}_{k}^{c}}' \right \|_{1}+\frac{\mu}{2} \left \| \boldsymbol{\alpha}_{k}^{c}-{\boldsymbol{\alpha}_{k}^{c}}'+\frac{\mathbf{z}_{k}}{\mu } \right \|_{F}^{2}
\end{equation}
Then ${\boldsymbol{\alpha}_{k}^{c}}'$ could be derived by operating $\mathrm{Threshold}(\boldsymbol{\alpha}_{k}^{c}+\frac{\mathbf{z}_{k}}{\mu },\frac{\lambda }{\mu })$. The operation of soft threshold is shown as follows.
\begin{equation}
\left [ \mathbf{S}_{\lambda /\mu }(\boldsymbol{\beta})  \right ]_{i}=\left\{\begin{matrix}
\hspace*{1cm} 0  \hspace*{2.6cm} \left | \boldsymbol{\beta}_{j} \right |\leq \lambda /\mu\\
\boldsymbol{\beta}_{i}-\mathrm{sign}(\boldsymbol{\beta}_{i})\lambda /\mu   \hspace*{1cm} \mathrm{otherwise}
\end{matrix}\right.
\end{equation}
where $\boldsymbol{\beta}_{i}$ means the value of the $i$-$th$ component of $\boldsymbol{\beta}$. After getting $\boldsymbol{\alpha}_{k}^{c}$ and ${\boldsymbol{\alpha}_{k}^{c}}'$, $\mathbf{z}_{k}$ and $\mu$ can be updated following $\mathbf{z}_{k}=\mathbf{z}_{k}+\mu(\boldsymbol{\alpha}_{k}^{c}-{\boldsymbol{\alpha}_{k}^{c}}')$ and $\mu=1.2\mu$.

\textbf{Update $\boldsymbol{\alpha}_{k}^{s}$:}
After acquiring ${\boldsymbol{\alpha}_{k}^{c}}$, the optimization of Eq. (9) can be reformulated to Eq. (15) at the step of updating $\boldsymbol{\alpha}_{k}^{s}$.
\begin{equation}
\boldsymbol{\alpha}_{k}^{s}=\arg\min \left \|  \mathbf{y}_{k}-\mathbf{D}_{k}(\boldsymbol{\alpha}_{k}^{c}+\boldsymbol{\alpha}_{k}^{s}) \right \|_{F}^{2}+\lambda \left \| \boldsymbol{\alpha}_{k}^{s} \right \|_{1}
\end{equation}
In fact, there are many methods to tackle the problem (13). For example, both ALM and Iterative Projection Method (IPM) [9] could deal with it. In this paper, we use IPM to address Eq. (13), as described in Algorithm 1.

\begin{algorithm}[!tbp]\small
\caption{Algorithm of updating $\boldsymbol{\alpha}_{k}^{s}$ in MMSSL}\label{alg1}
\begin{algorithmic}[1]
\renewcommand{\algorithmicrequire}{\textbf{Input:}}
\renewcommand{\algorithmicensure}{\textbf{End}}
\REQUIRE $\sigma$, $\gamma=\lambda/2$, $\mathbf{y}_{k}$, $\mathbf{D}_{k}$, and $\boldsymbol{\alpha}_{k}^{c}$, $k=1,\cdots , K$
\renewcommand{\algorithmicrequire}{\textbf{Initialization:}}
\renewcommand{\algorithmicensure}{\textbf{End}}
\REQUIRE $\tilde{\boldsymbol{\alpha }}_{k}^{s(1)}\mathbf{=0}$ and $h=1$,
\FOR{$k=1,...,K$}
\WHILE {not converged}

\STATE
h=h+1
\STATE
 $\tilde{\boldsymbol{\alpha }}_{k}^{s(1)}$=$\mathbf{S}_{\gamma/\sigma}\left ( \tilde{\boldsymbol{\alpha }}_{k}^{s(h-1)}-\frac{1}{\sigma}\triangledown \mathbf{F}(\tilde{\boldsymbol{\alpha }}_{k}^{s(h-1)}) \right )$\\
 where $\triangledown\mathbf{F}(\tilde{\boldsymbol{\alpha }}_{k}^{s(h-1)})$ is the derivative of the left of Eq. (15) $\left \| \mathbf{y}_{k}-\mathbf{D}_{k}(\boldsymbol{\alpha} _{k}^{c}+\boldsymbol{\alpha} _{k}^{s}) \right \|_{F}^{2}$, and $\mathbf{S}_{\gamma/\sigma}$ is a soft threshold operator that defined in Eq. (14);
 \ENDWHILE
\ENDFOR
\renewcommand{\algorithmicrequire}{\textbf{Output:}}
\renewcommand{\algorithmicensure}{\textbf{End}}
\REQUIRE $\boldsymbol{\alpha} _{k}^{s}=\tilde{\boldsymbol{\alpha }}_{k}^{s(h)}$, $k=1,\cdots , K$
\end{algorithmic}
\end{algorithm}

The complete algorithm of MMSSL is summarized in Algorithm 2. Note that, since we introduce a new variable $\boldsymbol{\alpha}_{k}^{c'}$ at the step of updating $\boldsymbol{\alpha}_{k}^{c}$, another reasonable way is that we could alternatively update $\boldsymbol{\alpha}_{k}^{c'}$ and $\boldsymbol{\alpha}_{k}^{c}$ until convergence, and then get the solution of $\boldsymbol{\alpha}_{k}^{c}$. At this time, the value of $\boldsymbol{\alpha}_{k}^{c}$ and $\boldsymbol{\alpha}_{k}^{c'}$ is similar.
\begin{algorithm}[!tbp]\small
\caption{Multi-Modal Similar and Special Learning (MMSSL)}\label{alg1}
\begin{algorithmic}[1]
\renewcommand{\algorithmicrequire}{\textbf{Input:}}
\renewcommand{\algorithmicensure}{\textbf{End}}
\REQUIRE $\lambda$, $\tau$, $\mathbf{y}_{k}$, $\mathbf{D}_{k}$, $k=1,\cdots , K$
\renewcommand{\algorithmicrequire}{\textbf{Initialization:}}
\renewcommand{\algorithmicensure}{\textbf{End}}
\REQUIRE $\boldsymbol{\alpha }_{k}^{c}\mathbf{=0}$, $\boldsymbol{\alpha }_{k}^{s}\mathbf{=0}$,
\WHILE {not converged }

\STATE
\textbf{Update coefficients $\boldsymbol{\alpha }_{k}^{c}$}: fix $\boldsymbol{\alpha }_{k}^{s}$\\
(a) compute $\boldsymbol{\alpha }_{k}^{c}$ following Eq. (12)\\
(b) compute ${\boldsymbol{\alpha }_{k}^{c}}'$ following Eq. (13)\\
(c) $\mathbf{z}_{k}=\mathbf{z}_{k}+\mu(\boldsymbol{\alpha }_{k}^{c}-{\boldsymbol{\alpha }_{k}^{c}}')$\\
\STATE
 \textbf{Update coefficients $\boldsymbol{\alpha }_{k}^{s}$}: fix $\boldsymbol{\alpha }_{k}^{c}$, and solve $\boldsymbol{\alpha }_{k}^{s}$ following \textbf{Algorithm 1}
 \ENDWHILE
\renewcommand{\algorithmicrequire}{\textbf{Output:}}
\renewcommand{\algorithmicensure}{\textbf{End}}
\REQUIRE $\boldsymbol{\alpha} _{k}^{c}$ and $\boldsymbol{\alpha} _{k}^{s}$ $k=1,\cdots , K$
\end{algorithmic}
\end{algorithm}

\subsection{The Classification Rule of MMSSL}
After obtaining the representation coefficients, the decision is ruled in favor of the class with total lowest reconstruction residual over all $K$ tasks.
\begin{equation}
j^{*}=\min \hspace*{0.2cm} \sum_{k=1}^{K}w_{k}\left \| \mathbf{y}_{k}-\mathbf{D}_{k,j}(\boldsymbol{\alpha}_{k,j}^{c}+\boldsymbol{\alpha}_{k,j}^{s}) \right \|_{F}^{2}
\end{equation}
where $\mathbf{D}_{k,j}$, $\boldsymbol{\alpha}_{k,j}^{c}$ and $\boldsymbol{\alpha}_{k,j}^{s}$ are the elements of the dictionary $\mathbf{D}_{k}$, the similar coefficient $\boldsymbol{\alpha}_{k}^{c}$ and the specific coefficient $\boldsymbol{\alpha}_{k}^{j}$ of $j$-$th$ category, respectively; $w_{k}$ is the weight value corresponding to the $k$-$th$ task, which could be obtained by exploiting the method mentioned in [8].

\section{Experimental Results}
In this section, we conduct two types of experiments. Healthy versus DM classification is first provided. Then we present the numerical results of Healthy versus IGR classification. In both experiments, KNN [29], SVM [27], [28], SRC [4], GSRC (group sprase) [26] which are general and effective classifiers are used for each individual modal classification.
Without loss of generality, we concatenate different features of different modalities as a single vector and refer it as tongue feature, facial feature or sublingual feature, respectively. Thus, the $K$ in Eq. (8) is equal to 3.

\subsection{Image Dataset}
The tongue, facial and sublingual sample database comprises 434 samples split into 192 Healthy samples, 198 DM samples and 114 IGR samples. Each sample has three different modal images including tongue image, facial image and sublingual image, respectively. All images were captured at the Guangdong Provincial TCM Hospital, Guangdong, China, from the early 2014 to the late 2015. Healthy samples were verified through a blood test and other examination. If indicators from these tests fall within a certain range (set by the Guangdong Provincial TCM Hospital), they were regarded as healthy. The FPG test was applied to diagnose whether a sample was suffering from the DM or IGR. When using the FPG test, all the samples had gone at least 12 hours without taking any food. For the DM patients, the blood glucose level was equal or larger than 7.11mmol/L, while the blood glucose level of IGR samples was between 6.1mmol/L and 7.11mmol/L. All these standard indictors are decided by the Guangdong Provincial TCM Hospital.

\subsection{Healthy Versus DM Classification}


\begin{figure*}
  \centering
  \includegraphics[width=180mm]{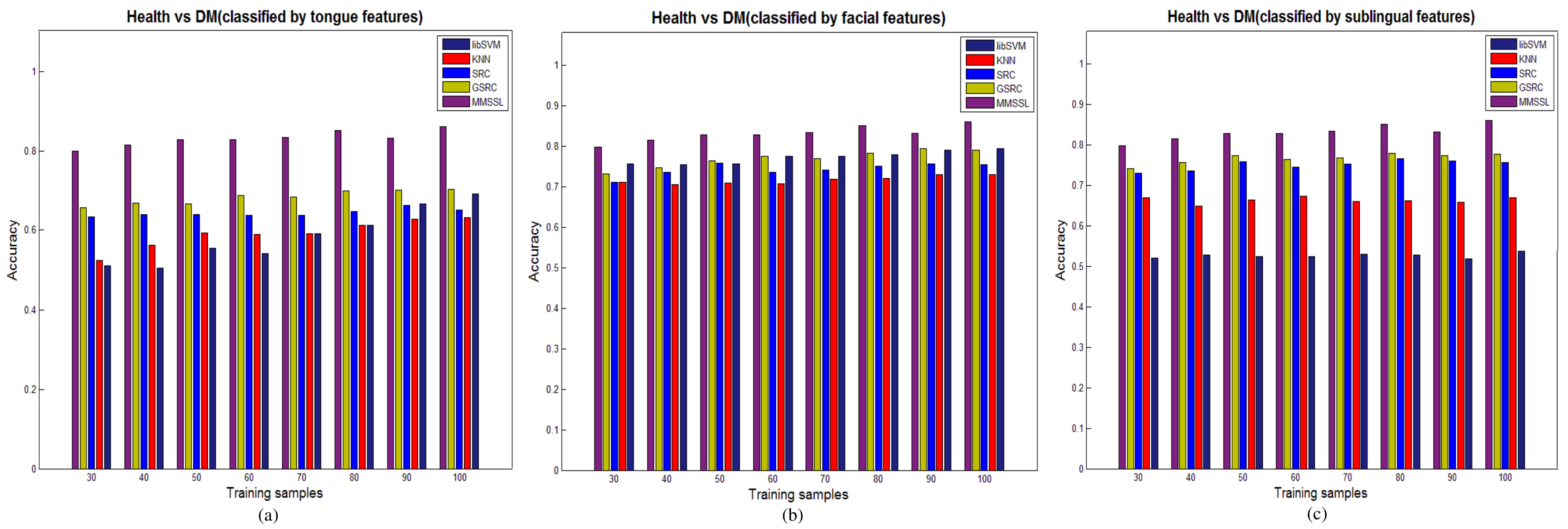}
  \caption{Comparison of Healthy and DM performance of single-modal and multi-modal classification methods. (a) Comparison of our method with the tongue image based feature. (b) Comparison of our method with the facial image based feature. (c) Comparison of our method with the sublingual image based feature.}
  \label{fig:env}
\end{figure*}

We first test the performance of our proposed multi-modal classification method in identification of DM for healthy controls, with the tongue, facial and sublingual tasks. We randomly select the number of training samples from 30 to 100, and the rest samples are used for testing. Fig. 4 illustrates the experimental results of our MMSSL approach, compared with the strategies using each individual modality with different classifiers. Note that Fig. 4 only shows the averaged results of 5 independent experiments. It is easy to see that the combined measurements of tongue, facial and sublingual features consistently achieve more accurate discrimination between DM patients and healthy controls. Particularly, compared with single tongue modal feature, our method MMSSL achieves about more than 15\% accuracy. The classification accuracies obtained MMSSL are gradually rising with the increasing number of training samples, and they are all higher than 80\%. In contrast, the best accuracy on face based modality is only close to 80\%.

In addition, Fig. 5 further plots the ROC curves of different classification methods for DM detection when the number of training samples is 70. Note that we only show the ROC curves of SRC and GSRC methods with the single modality to compare with our approach. As the output of SRC and GSRC is reconstruction errors. We use the ratio of different classes to represent the percentage. From Fig. 5, we can see that the area covered by MMSSL based ROC curve is obviously larger than other methods and modalities based curves.


\begin{figure}
  \centering
  \includegraphics[width=83mm]{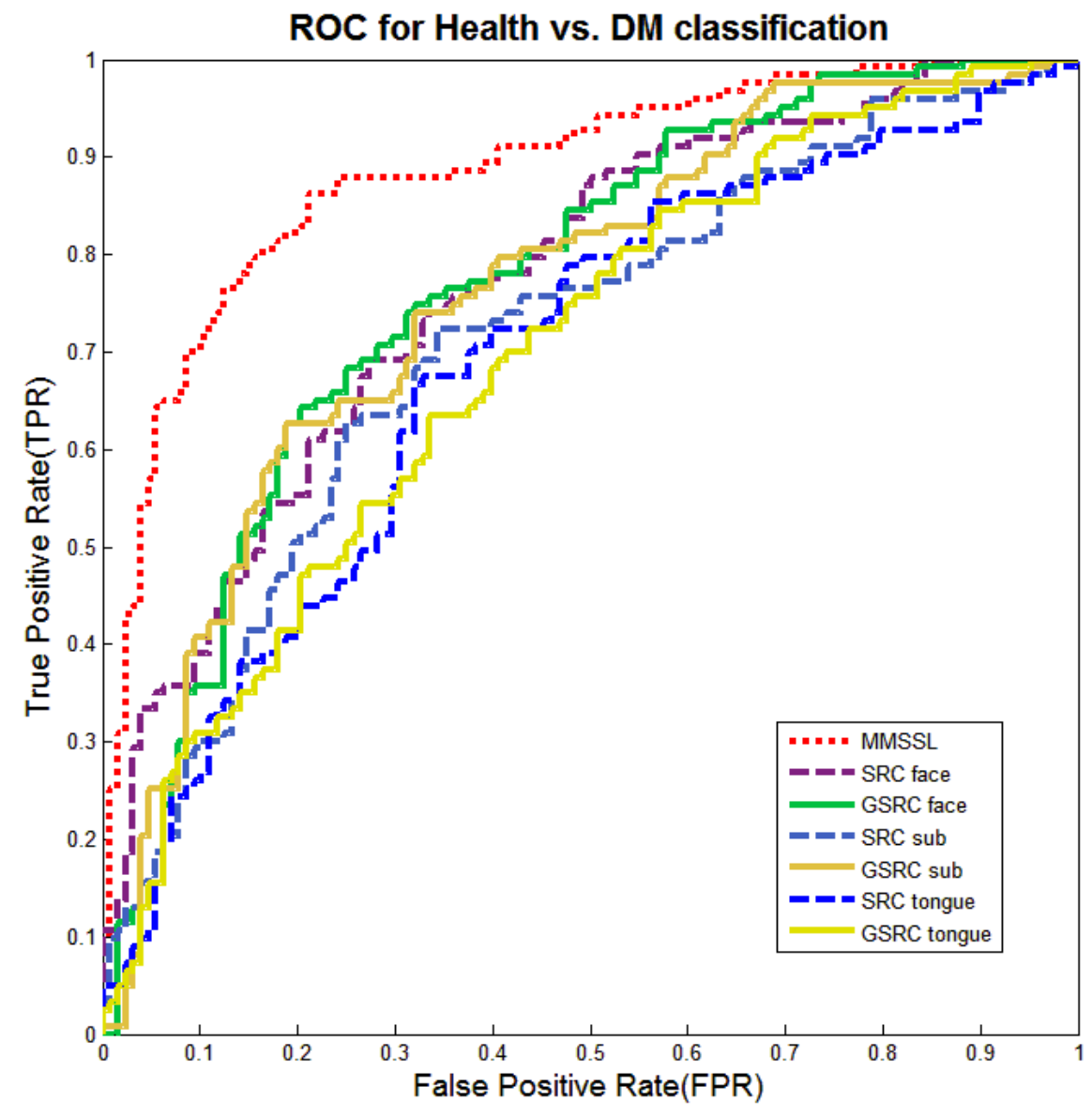}
  \caption{ROC curves of different methods and different features for DM classification.}
  \label{fig:env}
\end{figure}

Besides, we also illustrate the maximal and minimal classification rate in 5 independent experiments in Table 1. Note that we only make a comparison between MMSSL and the best performance on the single modality (face modality in this experiment). As we can observe from Table 1, the maximal and minimal rates increase with the increasing number of training samples. Compared with other methods, our proposed methods have a visible improvement both in maximal and minimal accuracies. Specially, when the number of the training samples reaches 100, the discrimination of maximization and minimization obtained by MMSSL achieve 89.01\% and 84.29\% respectively, while the best values acquired by other approaches are only 83.77\% and 75.39\%.
\begin{table*}\normalsize

\caption{The maximal and minimal classification rate in 5 independent experiments for DM detection.}
\centering
\begin{tabular}{|c |c| c |c| c| c| c| c| c| c|}
\hline
\multicolumn{2}{|c|}{}&\multicolumn{8}{c|}{Training samples}\\
\hline
{Methods} & {Value}&30&40 &50 &60 & 70 &80 &90 &100\\
\hline
\multirow{2}{*}{MMSSL }&{$\max$}&84.29\%&83.28\%&84.88\%&85.98\%&86.06\%&86.15\%&86.73\%&89.01\%\\
&$\min$&77.34\%&79.10\%&80.76\%&80.07\%&82.27\%&83.35\%&80.57\%&84.29\%\\
\hline
\multirow{2}{*}{K-NN (face) }&$\max$&75.23\%&74.59\%&74.23\%&74.91\%&75.30\%&78.79\%&76.78\%&75.92\%\\
&$\min$&68.88\%&64.31\%&64.60\%&67.53\%&68.92\%&68.40\%&69.67\%&69.63\%\\
\hline
\multirow{2}{*}{libSVM (face) }&$\max$&77.95\%&78.78\%&81.10\%&79.70\%&82.07\%&82.68\%&81.52\%&83.77\%\\
&$\min$&72.81\%&71.38\%&72.85\%&75.28\%&73.71\%&73.59\%&76.30\%&74.87\%\\
\hline
\multirow{2}{*}{SRC (face)}&$\max$&76.13\%&78.14\%&78.35\%&78.60\%&76.89\%&79.22\%&80.57\%&79.58\%\\
&$\min$&65.86\%&66.56\%&74.23\%&69.37\%&68.92\%&66.67\%&72.99\%&72.25\%\\
\hline
\multirow{2}{*}{GSRC (face) }&$\max$&74.92\%&78.78\%&79.04\%&81.55\%&79.28\%&82.25\%&81.99\%&82.72\%\\
&$\min$&71.60\%&70.10\%&73.20\%&74.17\%&73.71\%&69.26\%&75.83\%&75.39\%\\
\hline
\end{tabular}
\end{table*}

\begin{figure*}
  \centering
  \includegraphics[width=180mm]{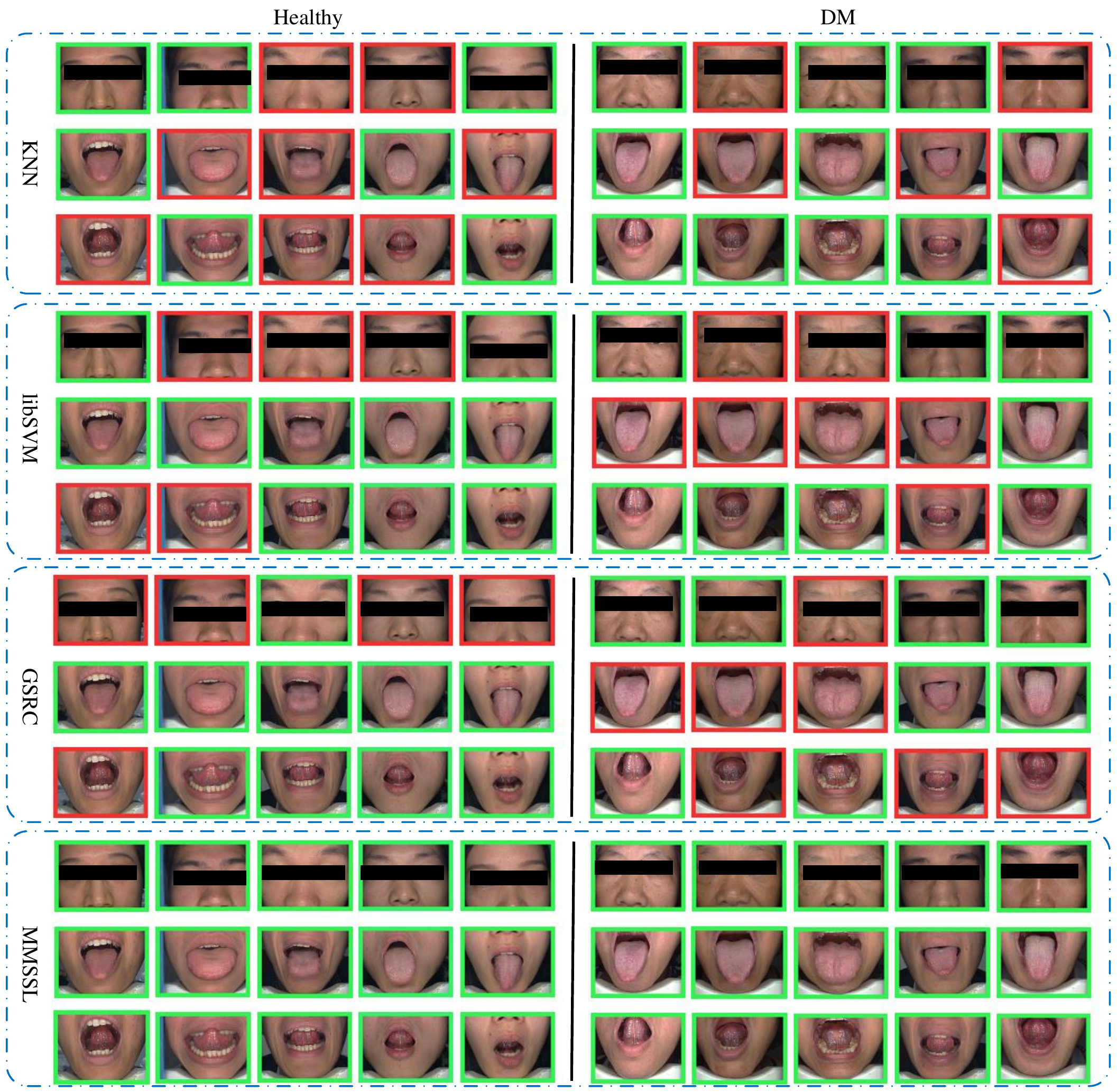}
  \caption{Example classification results of various classification methods on the individual modality for healthy and DM diagnosis. For each image, the red border indicates incorrect classification, and the green border indicates correct classification. KNN, SVM and GSRC fail in some images. In contrast, MMSSL has successfully classified all images. Particularly, MMSSL can also detect some samples that can not be classified by other methods with the individual modality.}
  \label{fig:env}
\end{figure*}

\begin{figure*}
  \centering
  \includegraphics[width=180mm]{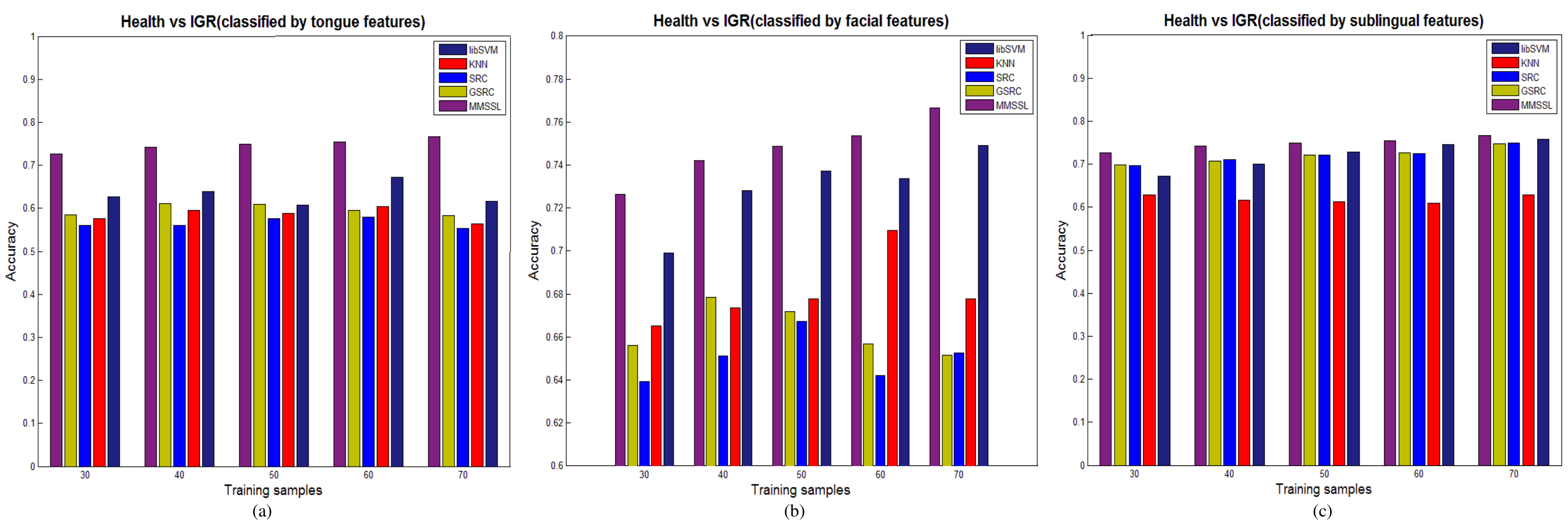}
  \caption{Comparison of Healthy Vs IGR performance of single-modal and multi-modal classification methods. (a) Comparison of our method with tongue image based feature. (b) Comparison of our method with facial image based feature. (c) Comparison of our method with sublingual image based feature.}
  \label{fig:env}
\end{figure*}

Some exemplar results of classification on DM and Healthy are shown in Fig. 6 (the corresponding exemplar results of SRC are shown in Fig. 1). As expected, SVM, SRC and GSRC are capable of discriminating healthy samples using tongue images, but fail to detect DM patients at many times. In contrast, SRC and GSRC have a prominent result on classifying DM using facial images, but fail to detect healthy samples. Different from aforementioned methods using an individual modality, our presented approach get an accurate result of each sample by jointly taking tongue, face and sublingual into account.

\subsection{Healthy Versus IGR Classification}



\begin{figure}
  \centering
  \includegraphics[width=83mm]{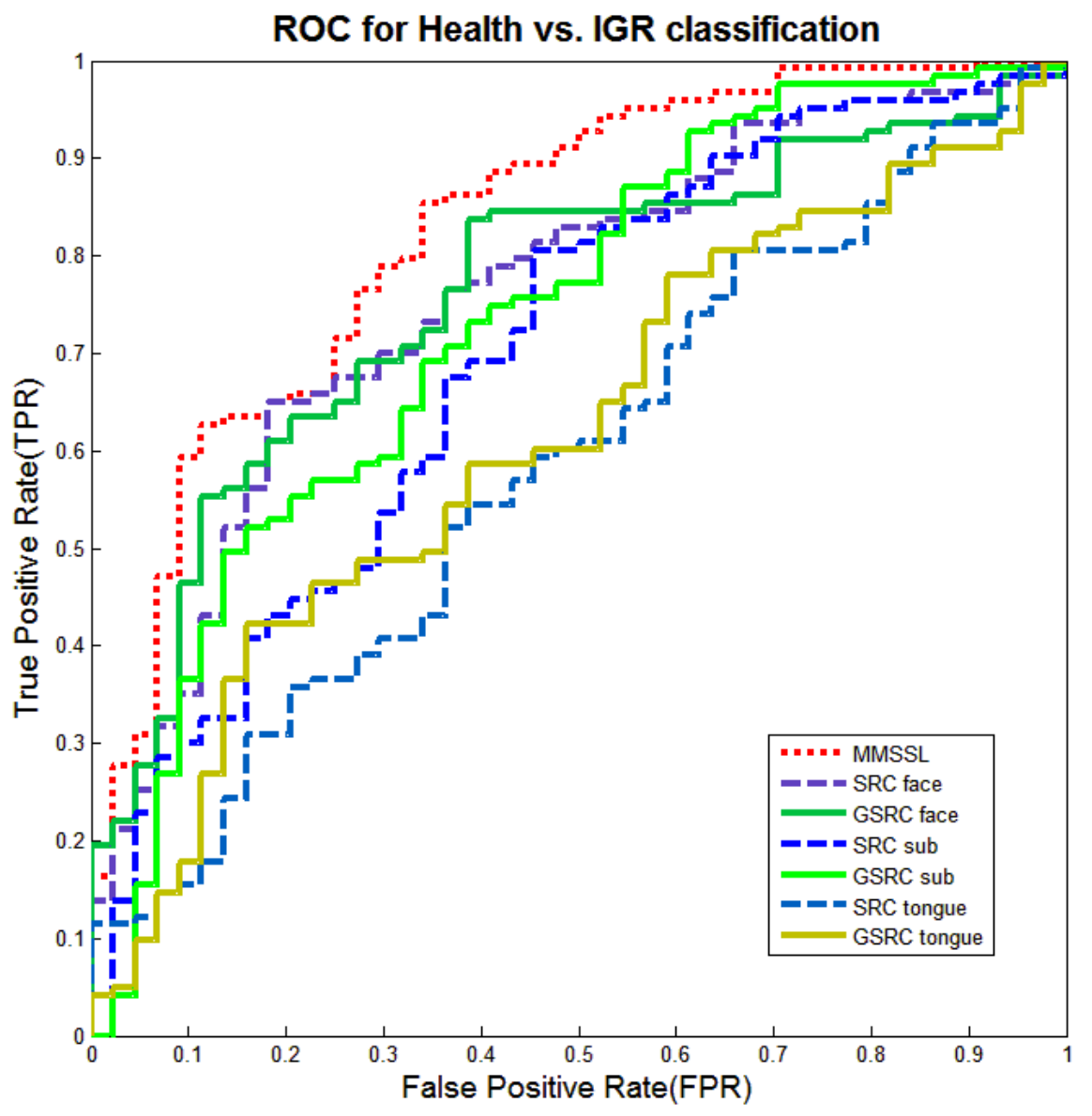}
  \caption{ROC curves of different methods and different features for IGR classification.}
  \label{fig:env}
\end{figure}

\begin{table*}\normalsize

\caption{The maximal and minimal classification rate in 5 independent experiments for IGR detection.}
\centering
\begin{tabular}{|c |c| c |c| c| c| c|}
\hline
\multicolumn{2}{|c|}{}&\multicolumn{5}{c|}{Training samples}\\
\hline
{Methods} & {Value}&30&40 &50 &60 & 70 \\
\hline
\multirow{2}{*}{MMSSL }&{$\max$}&75.30\%&77.97\%&77.29\%&79.14\%&81.44\%\\
&$\min$&70.85\%&70.48\%&73.91\%&73.80\%&73.05\%\\
\hline
\multirow{2}{*}{K-NN (sublingual) }&$\max$&71.90\%&67.52\%&69.76\%&66.84\%&69.46\%\\
&$\min$&64.35\%&60.77\%&60.14\%&57.75\%&57.08\%\\
\hline
\multirow{2}{*}{libSVM (sublingual) }&$\max$&68.83\%&72.69\%&73.91\%&77.01\%&77.84\%\\
&$\min$&66.40\%&68.28\%&71.50\%&72.19\%&74.25\%\\
\hline
\multirow{2}{*}{SRC (sublingual)}&$\max$&76.44\%&74.89\%&75.36\%&79.68\%&79.64\%\\
&$\min$&69.18\%&63.44\%&69.08\%&67.91\%&68.86\%\\
\hline
\multirow{2}{*}{GSRC (sublingual) }&$\max$&76.92\%&74.01\%&76.81\%&79.68\%&78.44\%\\
&$\min$&64.37\%&66.08\%&66.67\%&66.84\%&70.06\%\\
\hline
\end{tabular}
\end{table*}

In this subsection, we then apply our proposed multi-modal classification method in identification of IGR for healthy controls, with the tongue, facial and sublingual tasks, and make a comparison between MMSSL and other existing methods. Similar with the experiment on DM detection, the number of training samples from 30 to 70 are randomly selected with 5 times, and the rest samples are used for testing. Fig. 7 illustrates the averaged experimental results of our MMSSL approach, compared with the strategies using each individual modality with different classifiers. Our presented multi-modal classification strategy accomplishes a prominent rise in classification accuracy compared with other methods based on tongue, facial or sublingual features. For comparison with SVM, SRC and GSRC on sublingual features, the proposed method also get a slight enhancement in the averaged accuracy. Specially, the rate arrives at 76.68\% after combination with tongue, facial and sublingual tasks, while SVM with the sublingual task, which obtains the best result in all individual modalities, only performs 75.87\%.

The ROC curves of different classification methods for IGR diagnosis when the number of training samples reaches 70 is plotted in Fig. 8. Similarly, we only show the ROC curves of SRC and GSRC methods with the single modality to compare with our approach. The ROC curves demonstrate that the MMSSL has further performance compared with SRC (tongue), GSRC (tongue) and SRC(sublingual). Additionally, there is a slight improvement acquired by the MMSSL than that of GSRC (sublingual), GSRC (face) and SRC (face).

Table 2 illustrates the maximal and minimal classification rates in 5 independent experiments for IGR diagnosis. Similarly, only a comparison between MMSSL and the best performance on the single modality (sublingual modality in this experiment) is shown. Although at some times SVM have a better result than ours, our method carries out the best values of both maximization and minimization in most conditions. When the number of training samples is 70, the accurate discrimination is 81.44\% which is a slightly higher than that of SVM whose value is 79.64\%.

\section{Conclusion}
In this paper, a multi-modal fusion method for the Diabetes Mellitus and Impaired Glucose Regulation detection is proposed. The tongue, face and sublingual images are first captured by using a non-invasive capture device. Different features of these three types of images are then extracted. In order to exploit the correlation among them, we propose a novel fusion method to learn the common components and specific components of different modalities. Two types of experiments in identification of DM (or IGR) from healthy controls are conducted. The experimental results substantiate the effectiveness and superiority of our fusion method, compared with the case of using a single modality.


%



\section*{Acknowledgment}

The work is partially supported by the GRF fund from the HKSAR Government, the central fund from Hong Kong Polytechnic University, the NSFC fund (61332011, 61272292, 61271344) and Shenzhen Fundamental Research fund (JCYJ20150403161923528, JCYJ20140508160910917).

\ifCLASSOPTIONcaptionsoff
  \newpage
\fi


\begin{thebibliography}{1}

\bibitem{IEEEhowto:kopka}
World Health Organization, \emph{Prevention of blindness from diabetes mellitus},World Health Organization, 2006.

\bibitem{IEEEhowto:kopka}
Wang X, Zhang B, Yang Z, et al, \emph{Statistical analysis of tongue images for feature extraction and diagnostics},Image Processing, IEEE Transactions on, 2013, 22(12): 5336-5347.

\bibitem{IEEEhowto:kopka}
Zhang B, Kumar B V K, Zhang D, \emph{Noninvasive diabetes mellitus detection using facial block color with a sparse representation classifier}, Biomedical Engineering, IEEE Transactions on, 2014, 61(4): 1027-1033.

\bibitem{IEEEhowto:kopka}
Wright J, Yang A Y, Ganesh A, et al, \emph{Robust face recognition via sparse representation},Pattern Analysis and Machine Intelligence, IEEE Transactions on, 2009, 31(2): 210-227.

\bibitem{IEEEhowto:kopka}
Zhang B, Kumar B V K, Zhang D, \emph{Detecting diabetes mellitus and nonproliferative diabetic retinopathy using tongue color, texture, and geometry features},Biomedical Engineering, IEEE Transactions on, 2014, 61(2): 491-501.

\bibitem{IEEEhowto:kopka}
Lin Z, Chen M, Ma Y, \emph{The augmented lagrange multiplier method for exact recovery of corrupted low-rank matrices},arXiv preprint arXiv:1009.5055, 2010.

\bibitem{IEEEhowto:kopka}
Yang M, Zhang L, Zhang D, et al, \emph{Relaxed collaborative representation for pattern classification},Computer Vision and Pattern Recognition (CVPR), 2012 IEEE Conference on. IEEE, 2012: 2224-2231.

\bibitem{IEEEhowto:kopka}
Yuan X T, Liu X, Yan S, \emph{Visual classification with multitask joint sparse representation},Image Processing, IEEE Transactions on, 2012, 21(10): 4349-4360.

\bibitem{IEEEhowto:kopka}
Rosasco L, Verri A, Santoro M, et al, \emph{Iterative projection methods for structured sparsity regularization},2009.

\bibitem{IEEEhowto:kopka}
Shu T, Zhang B, \emph{Non-invasive Health Status Detection System Using Gabor Filters Based on Facial Block Texture Features},Journal of medical systems, 2015, 39(4): 1-8.

\bibitem{IEEEhowto:kopka}
Pang B, Zhang D, Li N, et al, \emph{Computerized tongue diagnosis based on Bayesian networks},Biomedical Engineering, IEEE Transactions on, 2004, 51(10): 1803-1810.
\bibitem{IEEEhowto:kopka}
Zhang D, Pang B, Li N, et al, \emph{Computerized diagnosis from tongue appearance using quantitative feature classification},The American Journal of Chinese Medicine, 2005, 33(06): 859-866.
\bibitem{IEEEhowto:kopka}
Zhang Y, Liang R, Wang Z, et al, \emph{Analysis of the color characteristics of tongue digital images from 884 physical examination cases},Journal of Beijing University of Traditional Chinese Medicine, 2005, 28(001): 73-75.
\bibitem{IEEEhowto:kopka}
Su W, Xu Z, Wang Z, et al, \emph{Objectified study on tongue images of patients with lung cancer of different syndromes},Chinese Journal of Integrative Medicine, 2011, 17: 272-276.

\bibitem{IEEEhowto:kopka}
Huang B, Wu J, Zhang D, et al, \emph{Tongue shape classification by geometric features},Information Sciences, 2010, 180(2): 312-324.

\bibitem{IEEEhowto:kopka}
Li B, Huang Q, Lu Y, et al, \emph{A method of classifying tongue colors for traditional chinese medicine diagnosis based on the CIELAB color space},Medical Biometrics. Springer Berlin Heidelberg, 2008: 153-159.

\bibitem{IEEEhowto:kopka}
Li C H, Yuen P C, \emph{Tongue image matching using color content},Pattern Recognition,2002, 35(2): 407-419.

\bibitem{IEEEhowto:kopka}
Kim B, Lee S, Cho D, et al, \emph{A proposal of heart diseases diagnosis method using analysis of face color},Advanced Language Processing and Web Information Technology, 2008. ALPIT'08. International Conference on. IEEE, 2008: 220-225.

\bibitem{IEEEhowto:kopka}
Liu M, Guo Z, \emph{Hepatitis diagnosis using facial color image},Medical Biometrics. Springer Berlin Heidelberg, 2008: 160-167.

\bibitem{IEEEhowto:kopka}
Wang X, Zhang D, \emph{An optimized tongue image color correction scheme},Information Technology in Biomedicine, IEEE Transactions on, 2010, 14(6): 1355-1364.

\bibitem{IEEEhowto:kopka}
Maciocia G, \emph{The foundations of Chinese medicine},Churchill Livingstone, 1989.

\bibitem{IEEEhowto:kopka}
Zhang H Z, Wang K Q, Jin X S, et al, \emph{SVR based color calibration for tongue image},Machine Learning and Cybernetics, 2005. Proceedings of 2005 International Conference on. IEEE, 2005, 8: 5065-5070.

\bibitem{IEEEhowto:kopka}
Zhang B, Wang X, Karray F, et al, \emph{Computerized facial diagnosis using both color and texture features},Information Sciences, 2013, 221: 49-59.

\bibitem{IEEEhowto:kopka}
Lin Z, Liu R, Su Z, \emph{Linearized alternating direction method with adaptive penalty for low-rank representation},Advances in neural information processing systems. 2011: 612-620.

\bibitem{IEEEhowto:kopka}
Chiu C C, Lan C Y, Chang Y H, \emph{Objective assessment of blood stasis using computerized inspection of sublingual veins},Computer methods and programs in biomedicine, 2002, 69(1): 1-12.

\bibitem{IEEEhowto:kopka}
Bengio S, Pereira F, Singer Y, et al, \emph{Group sparse coding},Advances in neural information processing systems. 2009: 82-89.

\bibitem{IEEEhowto:kopka}
Hsieh C J, Chang K W, Lin C J, et al, \emph{A dual coordinate descent method for large-scale linear SVM},Proceedings of the 25th international conference on Machine learning. ACM, 2008: 408-415.

\bibitem{IEEEhowto:kopka}
Fan R E, Chang K W, Hsieh C J, et al, \emph{LIBLINEAR: A library for large linear classification},The Journal of Machine Learning Research, 2008, 9: 1871-1874.

\bibitem{IEEEhowto:kopka}
Cover T M, Hart P E, \emph{Nearest neighbor pattern classification},Information Theory, IEEE Transactions on, 1967, 13(1): 21-27.

\bibitem{IEEEhowto:kopka}
Wang X, Zhang B, Guo Z, et al, \emph{Facial image medical analysis system using quantitative chromatic feature},Expert Systems with Applications, 2013, 40(9): 3738-3746.


\end{thebibliography}
\end{document}